\documentclass{Interspeech2024}
\usepackage{tipa}
\usepackage{adjustbox}
\usepackage{graphicx}
\usepackage{amsmath}
\usepackage{amssymb}
\usepackage{tikz}
\usetikzlibrary{shapes,arrows}
\usepackage{verbatim}
\usepackage{mathtools}
\usepackage{xcolor}

\usetikzlibrary{calc,positioning}

\interspeechcameraready

\title{Deciphering Assamese Vowel Harmony with Featural InfoWaveGAN}
\name[affiliation={1}]{Sneha}{Ray Barman}
\name[affiliation={1,2}]{Shakuntala}{Mahanta}
\name[affiliation={1,3}]{Neeraj Kumar}{Sharma}

\address{
  $^1$Centre for Linguistic Science \& Technology, $^2$Department of Humanities \& Social Sciences, 
  $^3$Mehta Family School of Data Science \& Artificial Intelligence\\
  Indian Institute of Technology Guwahati, Guwahati-781039, India}
\email{\{sneha.barman, smahanta, neerajs\}@iitg.ac.in}


\begin{document}
\maketitle

\begin{abstract}
\noindent Traditional approaches for understanding phonological learning have predominantly relied on curated text data. Although insightful, such approaches limit the knowledge captured in textual representations of the spoken language. To overcome this limitation, we investigate the potential of the Featural InfoWaveGAN model to learn iterative long-distance vowel harmony using raw speech data. We focus on Assamese, a language known for its phonologically regressive and word-bound vowel harmony. We demonstrate that the model is adept at grasping the intricacies of Assamese phonotactics, particularly iterative long-distance harmony with regressive directionality. It also produced non-iterative illicit forms resembling speech errors during human language acquisition. 
Our statistical analysis reveals a preference for a specific [+high,+ATR] vowel as a trigger across novel items, indicative of feature learning. More data and control could improve model proficiency, contrasting the universality of learning. 
\end{abstract}

\section{Introduction}
Understanding human language acquisition involves examining various aspects of grammar, with phonology playing a crucial role. Approaches such as 
Optimality Theory \cite{prince_optimality_2004} models phonological grammar as input-output pairs, selecting the most optimal output based on input. In contrast, Maximum Entropy Grammar \cite{hayes_maximum_2008} proposes that grammar constraints are language-specific and learned from data, employing machine learning techniques to determine these constraints.

Integrating neural network methodologies, multiple studies have recently proposed Generative Adversarial Network (GAN) and WaveGAN models for phonetic and phonological learning, with advancements like Featural InfoWaveGAN (fiwGAN) enhancing lexical learning and demonstrating proficiency in phonological representation learning across languages. A comprehensive review of prior generative linguistics studies is provided by Pater \cite{pater2019generative}. Begu\textipa{\v{s}} \cite{begus_generative_2020} proposed Generative Adversarial Network (GAN \cite{goodfellow_generative_2014}) as a method to model learnability more organically, wherein phonetic and phonological processes are represented as mappings from 
randomly distributed latent space to generative data. Donahue et al. \cite{donahue_adversarial_2018} introduced the WaveGAN model for audio data, employing Deep Convolutional GAN (DCGAN \cite{radford_unsupervised_2015}) architecture to learn language features from continuous speech signals. This model processes 1-second audio files sampled at $16$ kHz, converting them into vectors for input to the discriminator network. WaveGAN, adapted for phonetic and phonological learning \cite{begus_generative_2020}, effectively captures linguistic representations through a one-to-one mapping from latent space to generated data, mimicking how humans construct underlying phonological representations from speech streams. However, human acquisition also involves lexical learning, the ability to store unique information associated with meaning-bearing units or phonemes while adhering to language-specific phonotactics. Begu\textipa{\v{s}} \cite{begus_identity-based_2021} enhanced the WaveGAN architecture by integrating a lexical learning component, the Q-network \cite{chen_infogan_2016}, resulting in Featural InfoWaveGAN (fiwGAN). This advancement introduces a new latent space structure to comprehend featural representations of phonetic and phonological learning. The fiwGAN generator network learns to produce acoustic data by encoding unique lexical information and generating innovative acoustic data associated with distinct binary codes for each lexical item. Examining the fiwGAN network's performance on English allophonic distribution, Begu\textipa{\v{s}} \& Zhou \cite{begus_interpreting_2022} demonstrated its generative and learning capabilities by producing data comparable to human speech acquisition outputs. Furthermore, Chen \& Elsner \cite{chen_exploring_2023} evaluated FiwGAN's robustness by assessing English and French nasality, confirming its ability to learn phonological representations where latent variables correspond to identifiable phonological features. This highlights FiwGAN's proficiency as a phonological learner, which is particularly evident in controlled settings when comparing the same feature across languages.

Building on these studies, we investigate the potential of the Featural InfoWaveGAN model to learn iterative long-distance vowel harmony using raw speech data. Vowel harmony is a phonological phenomenon observed in many languages where vowels within a word tend to be influenced by one another in terms of their phonetic properties, such as tongue height, tongue position, or roundedness \cite{blevins2004evolutionary, rose2011harmony}. Owing to this phenomenon, certain vowels in a word may undergo changes to better match the phonetic characteristics of neighboring vowels. The result is an enhancement of the overall harmony and fluidity of pronunciation within the word \cite{ohala94b_icslp}. Vowel harmony can be categorized as regressive, where vowels are influenced by following vowels or progressive, where vowels are influenced by preceding vowels. We focus on Assamese, a language known for its phonologically regressive and word-bound vowel harmony \cite{mahanta_directionality_2008, archangeli_assamese_2020}. We demonstrate that the model is adept at grasping the intricacies of Assamese phonotactics, particularly iterative long-distance harmony with regressive directionality.

\section{Vowel Harmony in Assamese}
Assamese is an Indo-Aryan language spoken in the northeastern state of Assam (in India)by $15.3$ million individuals \cite{census_2011}. The variety analyzed in this paper is the colloquial Assamese spoken in the eastern region of Assam, also known as Upper Assam. This features eight surface vowels, namely, [i, e, \textepsilon, \textscripta, \textopeno, o, \textupsilon, u], and twenty consonants, namely [p, p\textsuperscript{h}, b, b\textsuperscript{h}, t, t\textsuperscript{h}, d, d\textsuperscript{h}, k, k\textsuperscript{h}, g, g\textsuperscript{h}, m, n, \textipa{\ng} s, z, x, h, \textturnr, j, w, l] \cite{mahanta_directionality_2008}.

\begin{table}[!ht]
\centering
\caption{Vowel inventory of Assamese}
\label{tab:accents}
\adjustbox{max width=0.45\textwidth}{
\begin{tabular}{llll}   
\hline
\textbf{Vowels} & \textbf{Front} & \textbf{Back} &  \textbf{ATR} \\
\hline
High & i &  u & +ATR \\
 &  &  \textupsilon& -ATR \\
Mid & e &  o & +ATR \\
 & \textepsilon & \textopeno& -ATR \\
Low &  &   \textscripta& -ATR \\ \hline
\end{tabular}}

\end{table}

\noindent 
In Assamese, the surface vowel occurrences are restricted in the following ways:
\textit{(a)} The two high vowels /i/ and /u/ are pronounced with an advanced tongue root [+ATR], as are the mid vowels /e/ and /o/; \textit{(b)} The mid vowels / \textepsilon/ and / \textopeno/ are slightly lower than /e/ and /o/ and are not realized with an advanced tongue root [-ATR], and \textit{(c)} The vowels /e/ and /o/ do not occur word-finally or initially, they occur only under circumstances of vowel harmony.

\begin{table} [!ht]
  \centering
  \caption{Examples of vowel harmony in Assamese \cite{mahanta_directionality_2008}}
\label{harmonytable}
  \resizebox{8cm}{!}{  
  \begin{tabular}{llllll}
    \hline
         \textbf{Stem} &  \textbf{Gloss} &  \textbf{-suffix} &  \textbf{Harmonised} & \textbf{Gloss} \\
         \hline 
          /p\textepsilon t/&  ‘belly’ &  -u &  [petu]& ‘pot-bellied’\\
         /p\textscripta g\textopeno l/ & `mad-M' & -i & [p\textscripta goli] & `mad-F'\\
         /g\textupsilon l/ &  `to mix' & -i & [guli] & `I mix'\\
         /b\textopeno s\textopeno r/ & `year' & -i & [bosori] & `yearly'\\
        \hline
\end{tabular}}
\end{table}

Assamese exhibits right-to-left ATR harmony where the high vowels /i/ and /u/ trigger [+ATR] harmony of [-ATR] high /\textupsilon/ and mid vowels [/\textepsilon/, /\textopeno/] in the language. From the examples of Table \ref{harmonytable}, we can argue that the stem vowels alternate under the influence of suffixes in Assamese. [+ATR] acts as a dominant feature that spreads from the suffix vowels in the leftward direction and affects vowels on the left side of the triggering vowel. As seen in the examples, harmony applies to all the [-ATR] vowels preceding a [+ATR] vowel within the word domain showcasing long-distance iterative harmony (see example 4 from Table 2).\cite{mahanta_directionality_2008} proposed that the harmony sequence in Assamese may be subject to the tendency of avoiding featural cooccurrence of [-ATR, +ATR] vowels. Therefore, [-ATR] vowels do not trigger harmony playing a recessive role. /\textscripta/ appears to be an opaque vowel that blocks harmony from its targeting vowels to its left (z\textupsilon n\textscripta ki `firefly' *zun\textscripta ki).  Exceptional patterns emerge when /\textscripta/ in the root morpheme is raised, followed by derivational suffixes [uw\textscripta] and [ij\textscripta]. /\textscripta/ is raised according to the preceding [-ATR] vowel; for example, when the preceding vowel is /\textepsilon/, /\textscripta/ becomes [e] (\textepsilon lah-elehuw\textscripta) and [o] otherwise. \cite{archangeli_assamese_2020} proposed that iterative harmony is gradient and vowels closer to the trigger is affected more than the ones further away from the trigger. Our study attempts to navigate these claims through modeling learnability computationally.

\section{Proposed Approach}
 The novelty of our study lies in computing iterative long-distance harmony from raw speech in an unsupervised manner. Vowel harmony is a complex phonological process requiring the learner to grasp crucial aspects like directionality, domains, features, iteration, locality, and opacity \cite{archangeli_harmony_2007}. The vowel harmony system of Assamese provides a crucial case study for directionality as it is neither stem-controlled nor dominant-recessive \cite{mahanta_directionality_2008}. We train the model with Assamese speech data containing both harmonic and non-harmonic words. We quantify the presence of [+ATR] harmony when the frequency of F1 of the target vowel [-ATR] is much lower in the vicinity of the trigger [+ATR] vowel than respective F2 and F3 \cite{mintz_infants_2018,olejarczuk_acoustic_2019}. Regressive directionality is anticipated when V2 explains the properties of V1 in a V1CV2 setting. We analyze the output audios statistically and observe that although the Generator strings vowels and consonants together to form a novel word, it does follow a vowel sequence within the word domain. We also identify that some innovative items have lexical meaning in Assamese, suggesting the emergence of lexical learning from the fiwGAN model.

\section{Model Architecture}
\begin{figure}
    \centering
    \includegraphics[width=8.5cm, height=6cm]{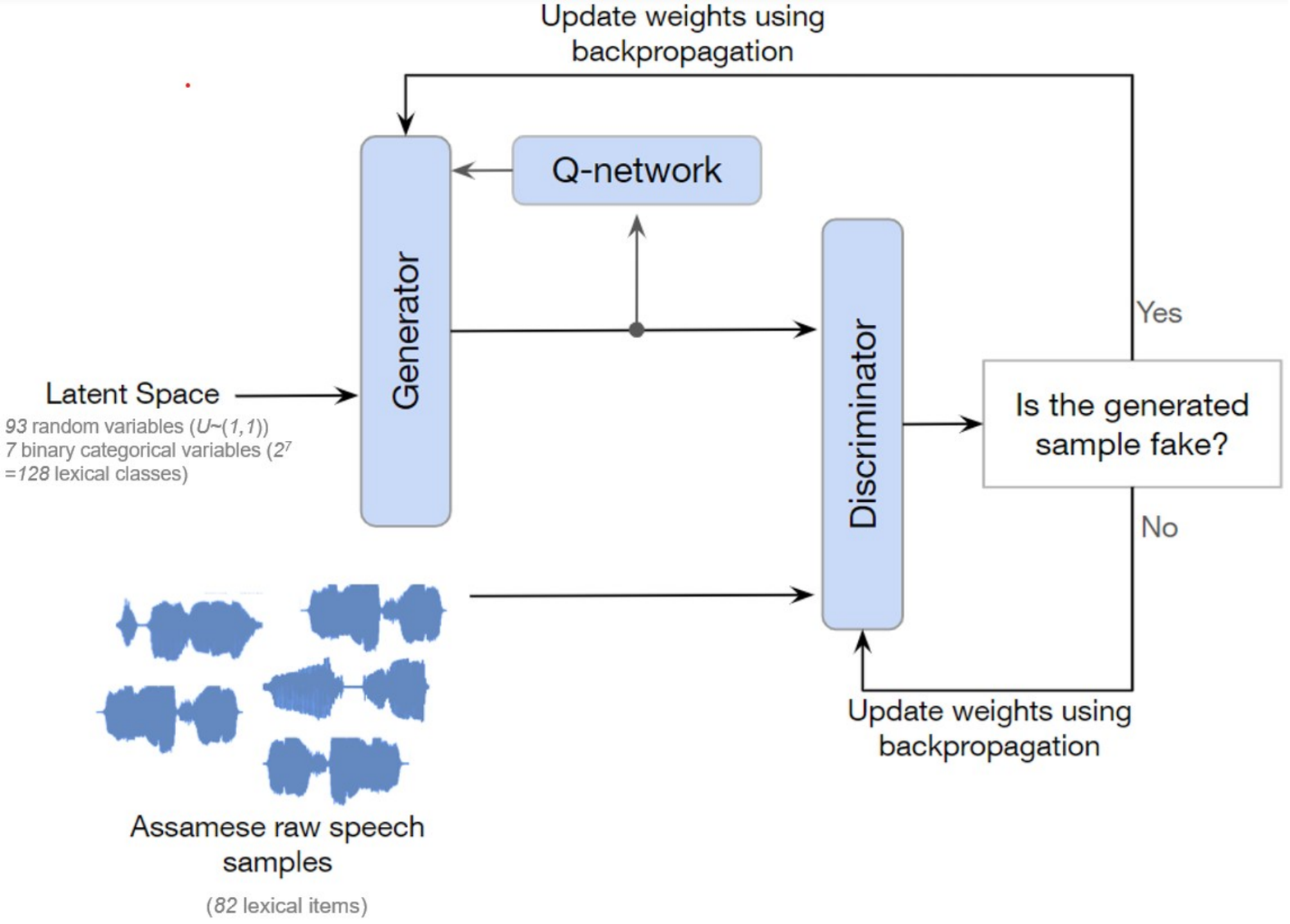}
    \caption{An illustration of fiwGAN architecture used in this work. We chose $N=82$ spoken utterances corresponding to lexical items in the Assamese language. The latent space contains 93 uniformly distributed latent variables, z, and 7 binary features ($\phi$) for 2\textsuperscript{7}=128 lexical classes.}
    \label{fig:GAN_block_diag}
\end{figure}

The fiwGAN neural network model comprises three primary components: a Generator, a Q-network, and a Discriminator (refer to Figure \ref{fig:GAN_block_diag}). Unlike traditional GAN models, where the Generator typically takes uniformly distributed latent variables as input, fiwGAN's Generator operates within a latent space composed of binary codes ($\phi$) and latent variables ($z$). Here, each binary variable ($\phi_n$) corresponds to a distinct feature \cite{begus_ciwgan_2021}. Trained to maximize the Discriminator's error rate while minimizing the Q-network's error, the Generator produces a vector containing approximately $16384$ data points for a $1$ second of audio data (sampled at $16$~kHz). These generated outputs, along with real acoustic data, are then assessed by the Discriminator, which computes the Wasserstein distance \cite{arjovsky_wasserstein_2017} between the generated and real data.

In parallel, the Q-network, which shares the same input data as the discriminator, operates differently. Its final layer comprises nodes corresponding to categorical variables ($\phi$). Unlike the Discriminator, the Q-network's loss function drives updates not only to its own weights but also to those of the Generator. This mechanism enables the Generator to associate each lexical item with a unique latent code, allowing the Q-network to retrieve the lexical information solely from the acoustic signals. This process facilitates lexical learning within the model. While the Generator produces raw acoustic data similar to but not an exact replication of the real input data, it effectively captures essential characteristics of the acoustic signals. \footnote{We used the Pytorch implementation of fiwGAN made available by \cite{begus_ciwgan_2021} at \url{https://github.com/gbegus/ciwganFiwGAN-pytorch.git}.}

\section{Materials}
\subsection{Assamese Speech Data}
For this experiment, $15$ native Assamese speakers ($7$ males and $8$ females) between the ages of $19-35$ were consulted. The speech data was recorded in the soundproof booth at the Phonetics and Phonology laboratory of the Indian Institute of Technology Guwahati (India) using a Tascam DR-100 MKII recorder. The created speech dataset consisted of $82$ spoken words(40 harmonic and 42 non-harmonic)\footnote{The dataset is available at \url{https://github.com/sneha2599/IS2024_fiwGAN-for-vowel-harmony.git}}. Each target word was in a carrier sentence written in Assamese script, that is, \textit{``moi X buli kolu''}, which corresponds to \text{``I say X''} in English. The participants were asked to utter each sentence at least four times. The collected speech data was then curated through manual segmentation using the audio-visual utility in PRAAT\cite{boersma_praat_2009} software. This resulted in $4789$ clean speech utterances (or tokens), each corresponding to an Assamese word. 

\begin{table}[!h]
\caption{Example dataset for training fiwGAN}
\label{tab:accents}
\centering
\adjustbox{max width=0.45\textwidth}{
\begin{tabular}{llll}
\toprule
\textbf{Stem}& \textbf{-suffix}& \textbf{Surface} & \textbf{Harmony type} \\
\midrule
dil\textepsilon& -i& dilei & harmonic\\ 
nokoril\textepsilon& -u& nokorileu & harmonic\\ 
g\textopeno r\textopeno m & -\textopeno -t & g\textopeno r\textopeno m\textopeno t& non-harmonic\\
b\textepsilon p\textscripta r & -i & b\textepsilon p\textscripta ri& non-harmonic\\
\bottomrule
\end{tabular}}

\end{table}

\subsection{Model Implementation}
The fiwGAN model takes 4789 (3169 harmonic and 1620 non-harmonic) 1-second-long unannotated single-channel audio files sampled at 16 kHz in raw waveforms. The latent space consists of 93 uniformly distributed latent variables $z$ ($U$$\sim$(-1,1)) and 7 binary variables, allowing for 2\textsuperscript{7}=128 unique classes, represented as [1,0,0,0,0,0,0 for word1], [0,1,0,0,0,0 for word2], etc. The Generator and Discriminator were trained with Adam optimizer while the Q-network was trained with RMSProp algorithm at a .0001 learning rate with a batch size of 64. It ran for 72 hours on an NVIDIA RTX A4500 GPU at the CLST lab of IIT Guwahati. The Generator data was analyzed following 960 training epochs ($\sim$ 44000 steps). Based on their intelligibility and noise quality, 64 out of 100 outputs were examined in our experiment after manual listening and segmentation in Praat\cite{boersma_praat_2009}. We extracted the first formants of the vowels in training and generated data at 10-time points using Praat and analyzed only the mean value to quantify the presence of ATR vowel harmony\cite{boersma_praat_2009}, followed by regression analysis in R \cite{r_core_team_r_2021} to assess the learnability of directionality and other discrete categories related to vowel harmony.  

\begin{figure*}[!t]
    \centering
    \input{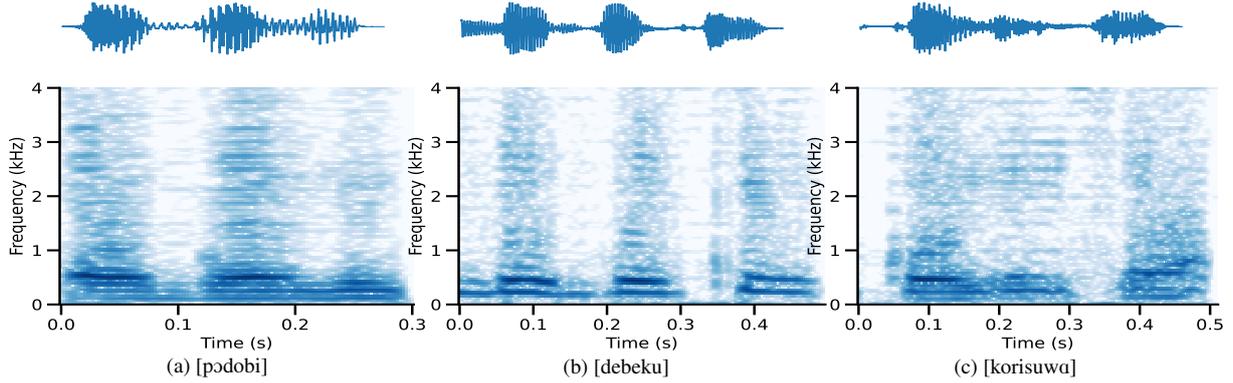}
    \caption{Spectrograms of three fiwGAN generated audio files: (a)[p\textopeno dobi], also an illicit item; (b) [debeku], also an innovative item following long-distance iterative harmony; and (c)[korisuw\textscripta] a novel word with lexical meaning in the Assamese language.}
    \label{fig:spectrogram}
\end{figure*}

\begin{figure}[!h]
    \centering
    \includegraphics[width=8.5cm, height= 5cm]{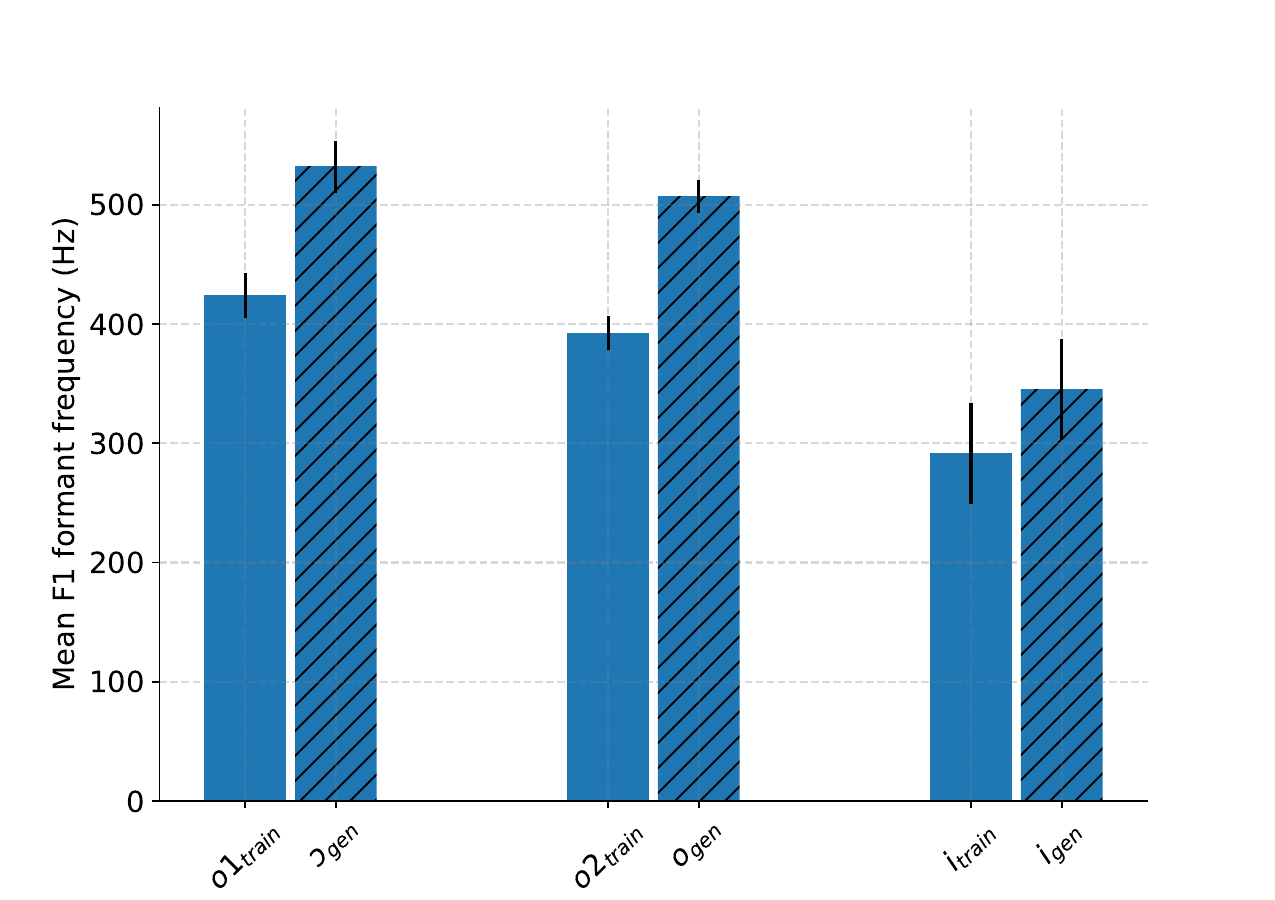}
    \caption{F1 comparison of [podobi] (training data; shown in bars) and [p\textopeno dobi] (generated data; shown in hatched bars). Here, o1 and o2 denote the first and second vowel, and i denotes the third vowel, in the input training data \textit{``podobi''}.}
    \label{fig:formants}
\end{figure}

\section{Results}
On the completion of each training epoch, the Generator produces $100$ audio files. In previous studies exploring English allophonic distributions and French nasality \cite{begus_identity-based_2021,chen_exploring_2023}, the Generator learned to produce speech-like sounds after close to $650$ epochs. In our case, the speech-like sounds were found to be generated after $720$ epochs. Intelligible sounds were generated after $800$ epochs. After $960$ epochs, the model started generating lexical items identical to the training dataset ([prohori],[\textepsilon k\textepsilon b\textscripta r\textepsilon],[p\textopeno l\textopeno x]) and innovative outputs ([dekhisi],[korobe],[korisuw\textscripta],[debeku] etc.). Additionally, some illicit forms ([n\textopeno korilu], [p\textopeno dobi], etc.) were also found in the outputs. An illustration of some outputs is shown in Figure~\ref{fig:spectrogram}.

Subsequently, we analyzed the innovative outputs and observed that the model strings together vowels and consonants from the training data; for example, [dekhisi] is assumed to be influenced by the word [dekhisu] in the input, while [korobe] may stem from [korobi]. In the context of vowel harmony, the novel items [debeku], [dekhisi], and [korisuw\textscripta] follow the vowel harmony pattern where [+high,+ATR] vowels [i,u] trigger harmony of the underlying [-ATR] vowels turning them into [+ATR]. We observed that [e]/[o] occurs when followed by [i]/[u] locally [korisuw\textscripta/dekhisi] as well as over a longer domain [debeku]. This suggests the capability of the model to learn the phonotactics of Assamese and the iterative long-distance vowel harmony.

Analyzing the illicit outputs, we observed that the vowels in them follow a specific pattern where the trigger vowels impact the immediately preceding vowel, suggesting non-iterative harmony([p\textopeno dobi] instead of [podobi]) (see Figure~\ref{fig:formants}). 
Moreover, [korisuw\textscripta] is close to a real-world compound word [kori suw\textscripta 'do see'] and a word the model has not encountered in the input, implying the emergence of lexical learning from the training.

\subsection{Statistical Analysis} We examined the generated outputs statistically to quantify the presence of discrete categories related to vowel harmony in Assamese 
The formant values of vowels in the training data were analyzed. It was observed that the first formant frequency (F1) of the [-ATR] vowels is much lower in the vicinity of the [+ATR] vowels (see Figure~\ref{fig:formants}).  We hypothesize that if V2 in the V1CV2 setting explains V1 better than V1 explains V2, the dataset follows regressive directionality. 
If not, the directionality is assumed to be left-to-right. A linear mixed-effects regression analysis was carried out on the training dataset \cite{bates_fitting_2015} (using R \cite{r_core_team_r_2021}). This helped examining the relationship between the dependent variable F1V1 and predictor variables V1 and V2 \footnote{V1full.model=lmer(F1V1$\sim$V1+V2+(1$\vert$word), data=formants, REML=FALSE)}. V1 and V2 are the fixed effects, while `word' is the random effect in the model. The likelihood ratio test compares the full model to a null model. The AIC score revealed a statistically significant difference between the two models (AIC(full)=1588.8, AIC(null)=1595.9). The chi-square test indicates a significant difference in model fit between the null and full models ($\chi\textsuperscript{2}$=33.062, df=13, p$<$0.001), reflecting a significant effect of V2 on the F1 value of V1 (see Table \ref{tab:lmer}). The dataset is then categorized into two subsets: harmonic and non-harmonic. Both the subsets are analyzed using the lmer model. The fixed-effects results from the full model for harmony\footnote{V1vhfull=lmer(F1V1$\sim$V1+V2+(1$\vert$word),data=har, REML=FALSE)} showed a distinction between V1, V2, and the intercept representing F1 of V1. The estimate of the intercept is about 552.129 Hz with a p-value$<$2e-16. A comparison of the full and null model using the likelihood ratio test showed that V2 affects V1 ($\chi\textsuperscript{2}$= 27.829, df = 7, p = 0.0002361), which reveals that the contrast is statistically significant (AIC(full)=467.04, AIC(null)=480.87)(see Table \ref{tab:lmer}). Assuming the results from statistical models for training data as a baseline, the generated items are annotated and then fit to linear regression models in R to examine the relationship between the dependent variable F1V1 and predictor variables V1 and F1V2 (see Table \ref{tab:lm}). The model was statistically significant (F(6, 15) = 9.504, p = 0.0002087), indicating that at least one predictor variable had a non-zero effect. The Adjusted R-squared value of 0.7084 suggests that the model explained approximately 70.84\% of the variability in F1V1. The coefficients for individual predictor variables were significant at various levels. Our regression model suggests that V1 and, to some extent, F1V2 are associated with variations in F1V1. Further analysis shows that the coefficients of V2[T.i] are significantly higher than other V2 variables (Estimate=-279.11, t-value=-3.376,p= 0.00817). This supports the idea that [i], a [+high,+ATR] vowel in Assamese, acts as a triggering vowel in the machine-generated outputs, suggesting the learnability of height as a trigger feature for Assamese vowel harmony. The overall analysis suggests that the model learns the right-to-left harmony pattern, which is evident in the influence of V2 on V1 for both first and second formant frequencies.

 


\begin{table}[!ht]
 \caption{LMER model for the training   dataset}
    \label{tab:lmer}
    \centering
        \resizebox{8cm}{!}{
    \begin{tabular}{cccccc}
    \toprule
      \textbf{Data} & \textbf{Directionality} &\textbf{Fixed effects} & \textbf{DF}  &\textbf{$\chi\textsuperscript{2}$} & \textbf{p} \\
         \midrule
       Whole & right-to-left &F1V1$\sim$V1+\textbf{V2}& 13  & 33.062 & $<$0.001 \\
        & left-to-right & F1V2$\sim$V2+\textbf{V1} & 10 &6.5156 & 0.77\\
        \hline
         Only [+ATR] & right-to-left & F1V1$\sim$V1+\textbf{V2}
& 7  & 27.829 & $<$ 0.001 \\
 & left-to-right & F1V2$\sim$V2+\textbf{V1} & 2 & 1.6522 & 0.43\\
    \bottomrule
    \end{tabular}}
\end{table}

\begin{table}[!h]
 \caption{Linear regression model for machine-generated items}
    \label{tab:lm}
    \centering
    \resizebox{8cm}{!}{    
    \begin{tabular}{cccc}
    \hline
        \textbf{Data} & \textbf{ Estimate} & \textbf{t-value} & \textbf{p-value} \\ \hline
       Whole & 605.25  & 7.793 & $<.001$\\
        only V2[i] coefficient &-279.11 & 3.376 & $.01$ \\ \hline
    \end{tabular}
    }
\end{table}
\section{Conclusion}

In summary, this study presented long-distance vowel pattern learning, emphasizing learners' acquisition of intricate vowel changes crucial for grasping vowel harmony. Unlike previous literature focusing solely on phonological learning, our findings mark an advancement, leveraging deep learning models to facilitate unsupervised learning of both acoustic and articulatory properties alongside underlying segmental features and vowel harmony. Through analysis of the Generator performance across epochs and manipulated latent space variables, insights into human-like speech generation have emerged. Our observations also indicate that, given limited data, the fiwGAN is an effective phonotactic learner, which is evident from its ability to learn complex systems like iterative long-distance harmony. Analysis of innovative items revealed instances of long-distance harmony alongside error items displaying non-iterative local harmony. The dominance of V2[i] as a trigger vowel indicates the model's capacity to learn trigger features, while statistical analysis suggests grammatical outputs following regressive directionality \footnote{\cite{dutta17_interspeech} observed coarticulation propensity in Assamese non-harmonic sequences which is strictly phonetic and not a goal of this paper.}. Moreover, lexical learning emerged post-training. Notably, no results with the opaque vowel [\textscripta] were observed, prompting further inquiry into the model's ability to identify vowel opacity.
\bibliographystyle{IEEEtran}
\bibliography{mybib}

\end{document}